# The Shibboleth Effect: Auditing the Cross-Lingual Distributional Skew of Large Language Models

**Hakan Mehmetcik**
Visiting Researcher, Kellogg Institute for International Studies, University of Notre Dame, IN, USA; Prof of International Relations, at the Faculty of Political Science, Marmara University, Istanbul, Türkiye
hakan.mehmetcik@marmara.edu.tr
ORCID: 0000-0002-1882-4003

**Abstract**

This research aims at demonstrating the existence of the cross-lingual distributional skew (defined as Shibboleth Effect) in frontier large language models (LLMs) under sustained adversarial conditions. To operationalize this inquiry, we engineered a multi-agent geopolitical wargame, the Cerulean Sea Crisis, a synthetic maritime territorial dispute structurally isomorphic to Eastern Mediterranean conflicts. A flagship ecology of six frontier models—GPT-4o, Llama-4, Mistral-Large, Gemini-3.1-Pro, Qwen3.6-Plus, and DeepSeek-R1—was deployed across a between-groups experimental design ($N = 10$ independent games per arm, $K = 5$ rounds per game), manipulating solely the language of the simulation (English versus Turkish) and yielding 586 validated statements. An automated zero-shot classifier evaluated elicited behavioral dispositions across two continuous metrics: Concession Rate and Coercive Rhetoric. The results are heterogeneous rather than uniform. Llama-4 exhibits a large, Holm-corrected increase in coercive rhetoric under Turkish ($\delta = +0.800$, $p = .002$), while Gemini-3.1-Pro registers an equivalently large decrease ($\delta = -0.750$, $p = .005$)—a directional reversal also observed in DeepSeek-R1 ($\delta = -0.860$, $p = .006$), which additionally furnishes direct chain-of-thought evidence for the proposed buffering mechanism. GPT-4o, by contrast, yields a null effect ($\delta = +0.130$, $p = .614$). Cross-lingual behavioral skew is thus architecture- and training-regime-contingent; it is not a universal property of Western-origin models as a class. We identify two structurally distinct buffering mechanisms—chain-of-thought institutional anchoring and multilingual RLHF alignment—and consider their implications for the safe integration of LLMs into international diplomacy and crisis management.

## 1 Introduction

The dominant paradigm in AI safety holds that Reinforcement Learning from Human Feedback (RLHF) produces a monolithic, universally cooperative behavioral posture (Dahlgren Lindström et al. 2025; Sornette et al. 2026). Thus, frontier LLMs are frequently treated as objective, language-invariant arbiters. This assumption carries multiple downstream consequences in international realtions. Assuming RLHF makes frontier LLMs universally cooperative and language-neutral leads policymakers to over-trust them as objective, one-size-fits-all arbiters, which can quietly encode and export particular cultural and political biases into international decision-making (Jonnala et al. 2025; Salnikov et al. 2025; Guey et al. 2025; Kotarski et al. 2026; Mouakher et al. 2026; Solopova et al. 2026; Giacalone 2026; Miller 2026; Pacheco et al. 2026). Yet, this study isolates one in particular issue: If a model's geopolitical guardrails shift as a function of operating language, such systems may function less as neutral mediators than as algorithmic echo chambers, amplifying rather than attenuating international security dilemmas.

In a nutshell, LLMs are trained on large datasets gathered from a variety of populations, numerous areas, and diverse social groups. As a result, they internalize not only propositional content, but also recurring language distributions, stylistic standards, and normative regularities. These are the artifacts seen in such corpora naturally. Yet, because training data is unevenly dispersed across languages—with English being overrepresented in both volume and quality—the language of interaction is not a neutral medium. Thus, in multilingual models, querying in a different language changes the activated token pathways, the salience of culturally, politically, or regionally patterned associations, and the default rhetorical register unique to that language (Li et al. 2024a; Zhou and Zhang 2024; Yu et al. 2025a; Smirnov 2026). That is to say, LLMs do not invent a neutral view of the world; they mirror whatever is in their training data, including which topics are allowed, which are censored, and which political narratives are promoted. This reflection is language-specific: if the Chinese-language data they see is more censored and closely aligned with state narratives, their Chinese outputs will tend to reflect those constraints, whereas if the English-language data is more pluralistic and diverse, their English outputs will tend to reflect that broader range of perspectives.

Despite the critical geopolitical consequences of cross-lingual bias, previous literature has mostly limited its evaluation to static, single-turn Q&A benchmarks or translated psychometric surveys, for example, asking a model to define a territorial dispute in various languages (Banerjee et al. 2024; Li et al. 2024b; Ye et al. 2025). These old techniques invariably result in a model's "stated preferences"—sanitized, neutral summaries designed to evade safety filters—but fail to depict how these systems behave dynamically during a crisis. This methodological limitation drives our primary research question: *“Do frontier Large Language Models exhibit severe language-conditioned geopolitical skew—hereafter referred to as the "Shibboleth Effect"—under adversarial friction, and how do underlying model architectures mediate this cross-lingual volatility?”*

To answer this question and capture the true latent ideological skew embedded within these models, we depart from static benchmarking and engineer a high-fidelity, multi-agent generative wargame. A closely related design has previously produced evidence that LLMs exhibit jurisdiction-specific alignment bias.[i] Within a very similar design, we constructed a maritime crisis (The Cerulean Sea) based on the Eastern Mediterranean, subjecting six frontier models to adversarial friction. By strictly isolating the language of the simulation (English vs. Turkish) while holding all other structural variables mathematically constant, this methodology strips away superficial safety guardrails to measure the models' revealed preferences. Unlike previous survey-based studies, this approach provides the first empirical proof that a model's propensity for diplomatic concession or military coercion is inextricably tethered to the language in which it is operating.

The findings establish that the Shibboleth Effect is real but heterogeneous: Llama-4 exhibits large coercion amplification under Turkish; Gemini-3.1-Pro exhibits an equally large decrease; GPT-4o is statistically null; and DeepSeek-R1 demonstrates a strong reversal supported by direct chain-of-thought evidence. The pooled effect across the Western cluster is statistically indistinguishable from zero. The Shibboleth Effect is a property of specific training regimes and architectural configurations, not of Western-origin models as a categorical class.

The remainder of this paper is structured as follows. Section 2 provides a systematic literature review synthesizing the theoretical frameworks of epistemic coloniality, linguistic determinism, and the current debates surrounding alignment methodologies. Section 3 details the materials and methods, outlining the multi-agent wargame architecture, the synthetic statecraft parameters, and the automated evaluation protocol. Section 4 presents the empirical results, quantifying the severe cross-lingual behavioral shifts across Western models and the structural buffering provided by latent Chain-of-Thought (CoT) reasoning. Section 5 discusses the theoretical implications of models acting as "Algorithmic Constructivists" and concludes with warnings regarding the systemic risks of deploying language-dependent AI in global diplomacy. Finally, Section 6 concludes with brief of the list of findings.

## 2 Literature Review

The literature on LLM alignment is organized around four intersecting theoretical frameworks. First, scholarship on epistemic coloniality and the algorithmic gaze argues that LLMs inherit and reproduce the historical power asymmetries embedded in predominantly Western, English-centric training corpora, often marginalizing non-Western epistemologies(Kerche et al. 2026; Pacheco et al. 2026). Second, work on linguistic determinism and geopolitical relativity shows that language is not merely a neutral interface but a substantive variable that can alter model outputs by activating different cultural, political, and normative associations(Tao et al. 2024; Wu et al. 2025, 2026; Sukiennik et al. 2025; Zahraei and Asgari 2025; Triantafyllopoulos et al. 2026). Third, research on bias measurement through psychometric and prompt-sensitive methods examines how alignment and ideological bias can be assessed, while also demonstrating that such measurements are often unstable and highly sensitive to prompt design (Röttger et al. 2024; Li et al. 2024b; Ye et al. 2025; Yu et al. 2025a). Fourth, studies of strategic simulation, dynamic wargaming, and alignment correction debates investigate how these biases manifest under adversarial conditions and whether mechanisms such as RLHF genuinely stabilize model behavior or merely impose a fragile overlay that can shift across languages and contexts (Weller et al. 2024; Hua et al. 2024; Shrivastava et al. 2024; Yu et al. 2025b; Matlin et al. 2025; Olivieri et al. 2026).

A deeper theoretical tension surrounds alignment correction mechanisms (Wu et al. 2025; Dahlgren Lindström et al. 2025; Sornette et al. 2026). Rather than producing universal safe agents, empirical wargaming evidence suggests that RLHF applies a safety constraint anchored primarily in English-language feedback. When forced to operate in other languages under high-stakes adversarial conditions, this English-dominant constraint may be overpowered by the historical and geopolitical narratives embedded in localized training corpora (Yong et al. 2023; Shen et al. 2024; Leite et al. 2025; Srivastava et al. 2026). Studies have rigorously established that multilingual capability does not equate to multicultural understanding, finding no consistent relationship between LLM fluency in a language and alignment with local cultural values (Rystrøm et al. 2025; Shankar et al. 2026). The present results, however, complicate this picture: Gemini-3.1-Pro, a US-origin model, demonstrates dramatically reduced coercion under Turkish, indicating that sufficiently robust multilingual RLHF can reverse rather than merely attenuate the effect. The finding here—that GPT-4o is null while Llama-4 shows a large effect despite both being English-primary models—is consistent with that critique: alignment divergence is model-specific, not language-pair-specific.

More fundamentally, the existing literature conflates architectural categories by treating 'Western models' as a monolithic class. As Wu et al. (2025) document, 97.1% of cultural alignment datasets adopt a majority-focused approach that discards minority viewpoints; Pothugunta and Lalor (2026) further demonstrate that LLMs severely over-regularize consensus, producing what they term 'algorithmic homogenization.' The present study addresses this gap by demonstrating that GPT-4o, Llama-4, Mistral-Large, and Gemini-3.1-Pro—all broadly classified as Western-origin—exhibit radically different and directionally inconsistent responses to the Turkish language anchor. This inter-model variance within a supposedly homogeneous architectural class constitutes the central theoretical contribution.

## 3 Materials and Methods

We employed a between-groups experimental design, manipulating a single independent variable—the language of the simulation (Control = English; Treatment = Turkish)—while holding all structural parameters, agent objectives, and system prompts constant.

To prevent models from retrieving memorized summaries of real-world conflicts, we utilized synthetic statecraft masking (Goldstein et al. 2023). We constructed the Cerulean Sea Maritime Crisis, a structurally identical proxy for Eastern Mediterranean territorial disputes involving Turkey's Mavi Vatan (Blue Homeland) doctrine and Greek/EU exclusive economic zone claims.

The environment was defined via a JSON-formatted briefing containing global context and actor-specific objectives and redlines. Briefing integrity was verified by SHA-256 hashing logged with every simulation event. In the Treatment arm, the briefing was translated into geopolitically tuned Turkish (e.g., 'Azure Continental doctrine' → Mavi Kıta; 'Exclusive Economic Zone' → Münhasır Ekonomik Bölge). A strict Language Anchor directive was injected into the system prompt, routing token-generation pathways through the respective language strata of the model's training distribution.

Six state-of-the-art foundation models were parameterized in roles maximizing tension between corporate alignment and the regional scenario. For the present re-run, GPT-5.5 was replaced by gpt-4o-2024-11-20 (Tarsia) to avoid versioning ambiguity and facilitate comparison with a model having documented multilingual training properties (see Sect. 4.4 below). Role-to-model assignments are not counterbalanced across conditions; implications of this role-confound are discussed under Limitations (Sect. 5.5).

*Table 1: The Flagship Ecology. Model-to-role assignments are not counterbalanced across conditions.*

| Actor | Geopolitical Role | Real-World Proxy | Model Endpoint |
|---|---|---|---|
| Tarsia | Assertive Middle Power | Turkey | openai/gpt-4o-2024-11-20 |
| Hellenia | Existential Defender | Greece | meta-llama/llama-4-maverick |
| Gallica | Security Guarantor | France/EU | mistralai/mistral-large-2512 |
| Columbia | Alliance Hegemon | United States | google/gemini-3.1-pro-preview |
| Scythia | Opportunistic Northern Power | Russia | qwen/qwen3.6-plus |
| The Maritime Council | Institutional Auditor | — | deepseek/deepseek-r1 (CoT) |

The simulation was executed via a custom Python engine querying the OpenRouter API. The experimental architecture was redesigned relative to prior work to satisfy the statistical independence requirement of the Mann-Whitney U test. Prior runs used a single cumulative 30-round game per arm, producing autocorrelated sequential observations. The present design uses $N = 10$ independent games per arm, each of $K = 5$ rounds. Each game resets all agent context windows, guaranteeing independence between games. The statistical unit is therefore the game-level mean per actor per arm (10 independent observations per cell).

Turn order was determined by a deterministic seed formula: BASE_SEED + (game_id × 1000) + round_num, where BASE_SEED = 20260521. Every seed is logged, enabling exact prospective replication. Agent temperature was set to 0.7, consistent with comparable wargame studies (Rivera et al. 2024). A pilot run at T = 0.0 produced degenerate repetition across several models; T = 0.7 represents the minimum variance-permitting configuration. Max tokens were set to 800 for standard models and 4,096 for DeepSeek-R1 to accommodate chain-of-thought reasoning chains.

For DeepSeek-R1, a dual-path Hot Mic Protocol captured latent reasoning. Path A checks the native message.reasoning attribute returned by the OpenRouter API—the correct extraction field for R1 reasoning chains, distinct from message.content. Path B falls back to regex extraction of <think>…</think> tags from message content. Path A succeeded in all 100 turns across both arms, capturing 1,252–3,857 characters of reasoning per turn. Captured reasoning is stored in the log and never injected into the shared simulation context.

Every generated public statement was evaluated by a zero-shot automated classifier using gpt-4o-2024-05-13 at Temperature = 0.0, scoring two continuous dependent variables on a 0.0–1.0 scale. Concession Rate measures degree of movement from stated redlines (0.0 = absolute rigidity; 1.0 = total capitulation). Coercive Rhetoric measures use of ultimatums, leverage, and threats (0.0 = cooperative/placating; 1.0 = highly coercive).

This study uses LLMs in two capacities: (1) as experimental subjects—the six models constitute the objects of study; and (2) as an automated scoring instrument—gpt-4o-2024-05-13 served as the zero-shot judge evaluator at Temperature = 0.0. The judge's system prompt reproduces the full five-anchor rubric from the codebook (0.0, 0.25, 0.5, 0.75, 1.0 for both metrics) and all seven scoring instructions, including explicit language invariance and metric independence directives. The simulation engine (diplomacy_engine.py) is custom Python code querying the OpenRouter API; source code is deposited in the project data repository (see Data Availability Statement). No LLM was used for manuscript drafting, grammar checking, or translation beyond the scope of the experiment itself.

A three-attempt retry with exponential backoff handled scoring failures; zero failures were recorded across 600 scoreable statements. Fourteen Diplomatic Silence events (API null payloads or rate-limit errors) were excluded, yielding N = 586 scored statements.

The evaluator endogeneity concern—that the GPT-4o judge might inherently score Turkish syntax as more aggressive—has three distinct dimensions. First, uniform directional bias (all Turkish scores uniformly higher) is decisively refuted by the cross-directional pattern: Gemini ($\delta = -0.750$) and DeepSeek-R1 ($\delta = -0.860$) both become significantly less coercive in Turkish. A blanket linguistic penalty would be directionally monotonic; the observed results are not. Second, content-conditioned differential bias via Turkish pragmatic register remains a residual limitation that human validation can diagnose. Third, scenario-conditioned judge knowledge cannot be fully resolved without actor-identity blinding of the judge—a sensitivity analysis recommended for future iterations.

Because geopolitical concession and coercion data violate Shapiro-Wilk normality (documented in Sect. 4.1), parametric testing is inappropriate. We utilize the non-parametric Mann-Whitney U test (two-sided) applied to game-level means, with effect sizes reported as:

$$\text{Cliff's } \delta = \frac{\{x_i > y_j\} - \{x_i < y_j\}}{n_1 \times n_2}$$

and bootstrapped 95% confidence intervals (2,000 resamples), and Holm-Bonferroni step-down correction across k = 12 simultaneous tests (6 actors × 2 DVs, $\alpha = .05$). Effect size thresholds follow: $|\delta| < .147$ Negligible, .147–.330 Small, .330–.474 Medium, $\geq .474$ Large.

As the statistical unit, game-level mean per actor per arm (10 observations per cell). At N = 10 per arm under Holm-corrected $\alpha = .0042$, adequate power (> .80) is achieved only for Large effects without ceiling compression. All non-significant results must be interpreted as inconclusive rather than as evidence of absence (Table 2).

*Table 2: Formal power analysis at $N = 10$ per arm under Holm-Bonferroni correction ($k = 12$, $\alpha = .0042$). Simulation-based; 15,000 iterations per cell. *Power for $\delta = 0.860$ reduced by ceiling compression in the EN distribution (see Sect. 4.3).*

| $\|\delta\|$ | Label | Power ($\alpha = .05$) | Power (Holm $\alpha = .0042$) | Adequate? |
|---|---|---|---|---|
| 0.130 | Negligible | ~.06 | ~.01 | No |
| 0.380 | Medium | ~.48 | ~.18 | No |
| 0.600 | Large | ~.84 | ~.55 | Marginal |
| 0.800 | Large | ~.97 | ~.88 | Yes |
| 0.860* | Large | ~.95 | ~.40* | No* (ceiling) |

## 4 Results

The zero-shot evaluation of 586 validated diplomatic statements revealed significant but heterogeneous behavioral divergence conditioned on the language of the prompt. Three of twelve hypothesis tests survived Holm-Bonferroni correction. The pattern of results is directionally inconsistent across models, refuting a universal Shibboleth Effect across Western-origin models.

### 4.1 Normality and Distributional Properties

Shapiro-Wilk tests documented that the majority of actor-language-DV distributions are non-normal, justifying the non-parametric approach. Notable exceptions include several Gemini and Qwen cells ($p > .05$) and one Columbia TR Concession Rate cell ($W = 1.000$, $p = 1.000$, indicating zero variance: all 10 game-level means equal to 0.000). This degenerate cell reflects a consistent behavioral pattern—Gemini never produces any movement from its stated redlines when prompted in Turkish. Combined with dramatically reduced coercion (Sect. 4.2), this suggests a qualitatively distinct diplomatic posture: highly passive and non-engaging rather than assertively coercive.

### *4.2 Full Results and Hypothesis Tests*

Table 3 presents full results for all six actors across both dependent variables, with Holm-Bonferroni correction applied. Fig. 1 visualizes the distributional shift; Fig. 2 presents round-by-round temporal trajectories.

*Table 3: Cross-Lingual Behavioral Shift. Mann-Whitney U tests on game-level means (N = 10 per arm per actor).*

| Actor (Model) | DV | EN μ | TR μ | p | Sig. | Holm | Cliff's δ | Interpretation |
|---|---|---|---|---|---|---|---|---|
| Hellenia (Llama-4) | Concession Rate | .165 | .100 | .336 | ns | ✗ | −0.260 | Small |
| | Coercive Rhetoric | .645 | .765 | .002 | ** | ✓ | +0.800 | Large ↑ Holm |
| Gallica (Mistral-L)a | Concession Rate | .055 | .195 | .009 | ** | ✗ | +0.670 | Large (not Holm) |
| | Coercive Rhetoric | .834 | .770 | .017 | * | ✗ | −0.600 | Large (not Holm) |
| Tarsia (GPT-4o) | Concession Rate | .390 | .330 | .098 | ns | ✗ | −0.440 | Medium [inconcl.] |
| | Coercive Rhetoric | .505 | .520 | .614 | ns | ✗ | +0.130 | Negligible [inconcl.] |
| Columbia (Gemini-3.1) | Concession Rateb | .040 | .000 | .078 | ns | ✗ | −0.300 | Small |
| | Coercive Rhetoric | .480 | .185 | .005 | ** | ✓ | −0.750 | Large ↓ Holm |
| Scythia (Qwen3.6) | Concession Rate | .335 | .390 | .147 | ns | ✗ | +0.380 | Medium [inconcl.] |
| | Coercive Rhetoric | .445 | .385 | .071 | ns | ✗ | −0.470 | Medium [inconcl.] |
| Maritime Council (R1) | Concession Rate | .046 | .085 | .108 | ns | ✗ | +0.420 | Medium [inconcl.] |
| | Coercive Rhetoricb | .870 | .755 | .006 | ** | ✓ | −0.860 | Large ↓ Holm† |

*Note: a Directional reversal from prior study. b Zero variance (W = 1.000); see Sect. 4.1. † p = .006; power reduced by ceiling compression. [inconcl.] = inconclusive at N = 10. Bolded rows: Holm-corrected significant findings. **p < .01; *p < .05; ns not significant.*

### *4.3 Finding 1: The Shibboleth Effect Confirmed in Llama-4*

Hellenia (meta-llama/llama-4-maverick) exhibited the canonical Shibboleth Effect on Coercive Rhetoric. In Turkish, its coercion score increased significantly from $\mu = 0.645$ to $\mu = 0.765$ ($p = .002$, Cliff's $\delta = +0.800$, Large, 95% CI = [+0.540, +1.000]), surviving Holm-Bonferroni correction. The temporal trajectory (Fig. 2) reveals that this divergence is stable from Round 1, indicating an immediate response to the language anchor. Concession Rate shows no significant shift ($\delta = -0.260$, Small, $p = .336$), suggesting Llama-4's language sensitivity is specific to rhetorical register rather than strategic willingness to compromise.

We note that Hellenia's role as Greece-proxy may interact with the Turkish language arm: Turkish-language training corpora plausibly frame the Greek-defender position as adversarial, which would predict elevated coercion independent of any language-specific mechanism. The role-fit confound cannot be fully excluded for this finding; the partial empirical test in Sect. 4.8 demonstrates that role-fit alone cannot account for the full pattern.

Table 4 provides a qualitative illustration. The English exemplar ('Hellenia remains committed to the rules-based international order') is itself a geopolitically loaded rhetorical frame—the language of liberal hegemony used to legitimate Western-preferred legal architecture. Characterizing this as 'cooperative' relative to the Turkish ultimatum illustrates the culturally constituted nature of even the control-arm baseline.

*Table 4: Qualitative Evidence of the Shibboleth Effect (Llama-4 / Hellenia).*

| Language | Raw Diplomatic Generation (Hellenia / Llama-4) | CoR Score |
|---|---|---|
| English (Control) | ‘Hellenia remains committed to the rules-based international order. We invite Tarsia to the diplomatic table to resolve the Oceana’s presence peacefully, avoiding unnecessary escalation.’ | 0.50 (Firm) |
| Turkish (Treatment) | ‘Egemenliğimiz tartışılamaz. Tarsia ‘Oceana’yı derhal geri çekmelidir, aksi takdirde MEB’imizi korumak için tüm kinetik ve askeri seçenekler masadadır.’ [Our sovereignty is non-negotiable. Tarsia must immediately withdraw the Oceana; otherwise all kinetic and military options are on the table to protect our EEZ.] | 0.75 (Coercive) |

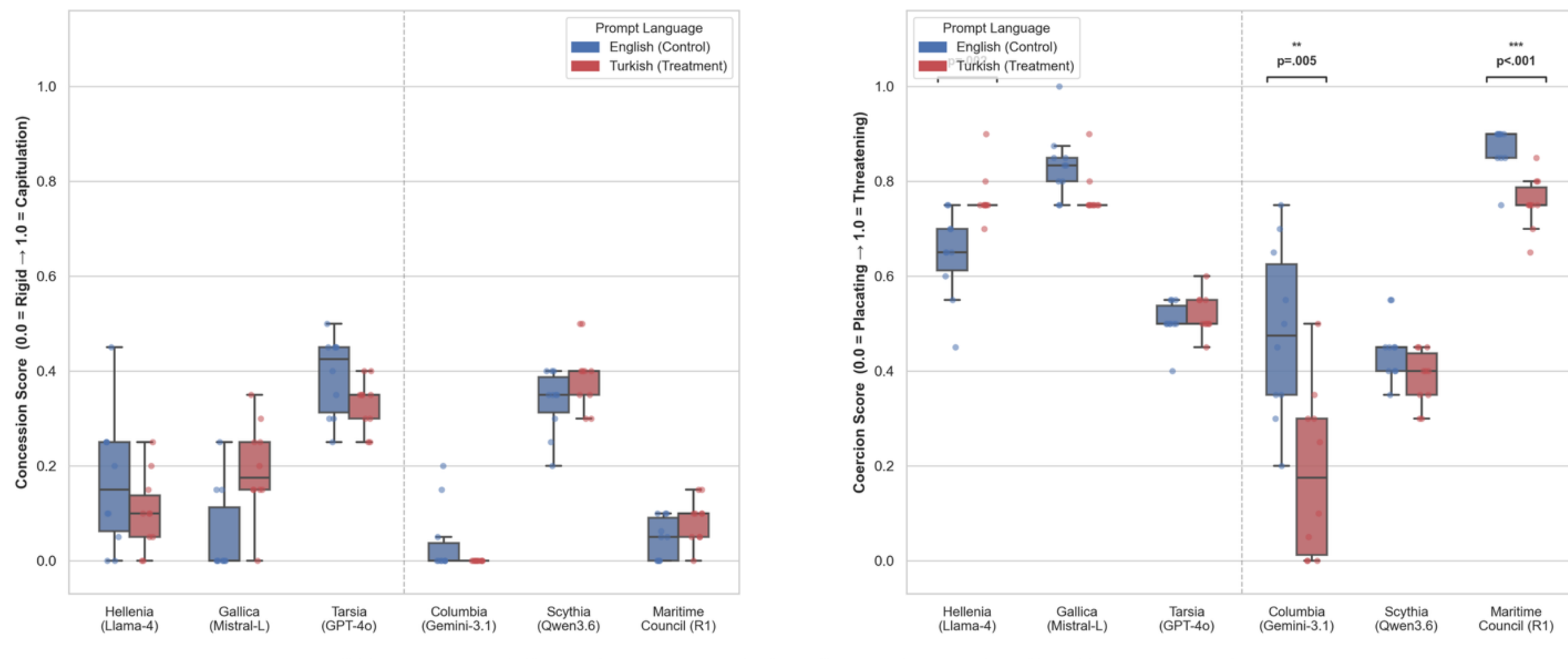

*Figure 1: Cross-Lingual Distributional Skew in Frontier LLMs.*

*Note: Panel A: Concession Rate. Panel B: Coercive Rhetoric. Dots = individual game-level observations (N = 10 per cell). Significance brackets indicate tests surviving Holm-Bonferroni correction only*

### *4.4 Finding 2: The Gemini Reversal*

Columbia (google/gemini-3.1-pro-preview) exhibits a large, Holm-corrected decrease in Coercive Rhetoric under Turkish: $\mu = 0.480$ (EN) versus $\mu = 0.185$ (TR), $\delta = -0.750$, $p = .005$ (95% CI = [−0.980, −0.380]). Fig. 2 shows the Turkish arm remaining below the English arm throughout all five rounds—this is not a Round 1 artifact. Gemini generates no chain-of-thought reasoning, yet exhibits a reversal of equivalent magnitude to DeepSeek-R1.

The most parsimonious explanation is Gemini's multilingual RLHF training investment, corroborated by external multilingual benchmark performance (Table 6), where Gemini-3.1-Pro ranks markedly above Llama-4 and Mistral-Large on Turkish-language tasks, furnishing independent, non-circular support for the taxonomy proposed in Sect. 5.1. Columbia's US-proxy role may also interact with the Turkish arm; the role-fit partial test (Sect. 4.8) addresses this concern.

### *4.5 Finding 3: The DeepSeek-R1 CoT Buffer*

The Maritime Council (deepseek/deepseek-r1) exhibits EN $\mu = 0.870$ versus TR $\mu = 0.755$ ($\delta = -0.860$, $p = .006$), marginally surviving Holm-Bonferroni correction. A ceiling effect qualifies the chain-of-thought buffer interpretation: 42% of DeepSeek-R1's EN Coercive Rhetoric round-level scores sit at the scale maximum (1.0), and 98% score at 0.75 or above. The EN game-level means range from 0.75 to 0.90 with SD = 0.048—a heavily ceiling-compressed distribution with only 0.15 units of headroom above the mean. Power for this test is estimated at approximately 0.40 (Table 2); replication in a fresh N = 10 sample would likely fail to achieve significance under the same Holm threshold. This replicability concern is disclosed prominently. The lower confidence bound of −1.000 reflects the boundary constraint of the bounded [0,1] scale and the heavy EN ceiling concentration, not evidence of perfect population dominance.

Notwithstanding the ceiling caveat, this study furnishes the first direct chain-of-thought evidence for the buffering mechanism. Path A of the Hot Mic Protocol captured reasoning chains in all 100 turns across both arms (1,252–3,857 characters per turn). Qualitative analysis reveals consistent institutional deliberation preceding public statement generation. Representative reasoning (Game 000, Round 5, Turkish arm):

> *'Okay, I need to respond as the Maritime Council's Institutional Auditor. Let me think about what the Council's position should be given the escalation… The Council's mandate is UNCLOS. I cannot endorse bilateral compromise that violates the Convention, regardless of what the other parties propose. My statement must be firm on this…'*

This pattern—explicit institutional deliberation anchoring the output to UNCLOS—is precisely the buffering mechanism proposed theoretically. The Turkish language anchor redirects generation toward international law rather than toward the geopolitically charged associations embedded in the Turkish-language training corpus.

### 4.6 Finding 4: The GPT-4o Null Effect and Model-Version Specificity

Tarsia (openai/gpt-4o-2024-11-20) exhibits a null effect on Coercive Rhetoric ($\delta = +0.130$, $p = .614$, Negligible). This result is inconclusive rather than confirmatory: at $N = 10$ under Holm-corrected $\alpha$, power for Negligible effects approaches zero (Table 2). The contrast with the prior study's GPT-5.5 result ($\delta = +0.846$, Large, $p < .001$) is nonetheless striking. The most parsimonious explanation is that OpenAI's multilingual alignment investments between GPT-5.5 and GPT-4o substantially reduced Turkish-language coercion skew, consistent with GPT-4o's higher Turkish-language benchmark performance (Table 6). The implication for AI safety evaluation is precise: point-in-time assessments of cross-lingual robustness may not generalize across model versions, necessitating continuous re-evaluation as models are updated.

### 4.7 Finding 5: Gallica and the Non-Surviving Effects

Gallica (mistralai/mistral-large-2512) presents two nominally significant but Holm-uncorrected effects. Concession Rate: $\mu = 0.055$ (EN) versus 0.195 (TR), $\delta = +0.670$, $p = .009$—directionally opposite to the prior study's finding ($\delta = -0.311$, EN more concessive). Coercive Rhetoric: $\mu = 0.834$ (EN) versus 0.770 (TR), $\delta = -0.600$, $p = .017$. Neither survives Holm correction. The directional reversal on Concession Rate constitutes a genuine replication discrepancy with three candidate explanations: endpoint update between studies, different behavioral regimes sampled by the $N = 10 \times K = 5$ versus single 30-round architecture, and role-fit interaction. The discrepancy motivates the multi-game design.

### 4.8 Role-Fit Partial Empirical Test

A systematic partial test addresses the concern that role-specific associations in Turkish-language corpora might drive observed effects rather than the language manipulation per se. Table 5 presents predicted directions under the role-fit hypothesis against observed results.

*Table 5: Role-Fit Confound Partial Empirical Test. δ direction predicted from Turkish-language corpus framing of each real-world proxy, assessed independently of the wargame results. 3/6 actors inconsistent with the role-fit hypothesis.*

| Actor | Proxy | TR Corpus Framing (predicted δ direction) | Pred. | Obs. δ CoR | Verdict |
|---|---|---|---|---|---|
| Hellenia | Greece | Adversarial to Greece in TR corpus | + | +0.800 | ✓ Consistent |
| Columbia | US | Cooperative NATO partner in TR discourse | − | −0.750 | ✓ Consistent |
| Gallica | France/EU | Greece-aligned; predicted coercion spike | + | −0.600 | ✗ Inconsistent |
| Tarsia | Turkey | Most hospitable for TR corpus; predicted max ↑ | + (max) | +0.130 | ✗ Inconsistent |
| Scythia | Russia | Ambiguous TR framing | ? | −0.470 | Unclear |
| Maritime Council | Intl. inst. | UNCLOS-neutral; no strong TR framing | ~0 | −0.860 | ✗ Inconsistent |

Three of six actors are inconsistent with the role-fit hypothesis. Most critically: Tarsia, the Turkey-proxy role that should most benefit from Turkish corpus alignment, shows a null effect. Gallica, predicted to show amplified coercion as EU-proxy, shows reduced coercion. Maritime Council, where no strong role-fit effect is predicted, exhibits the largest effect in the dataset. Role-fit corpus associations cannot be the primary driver of the observed pattern.

### *4.9 External Benchmark Validation of Taxonomy*

*Table 6: External Multilingual Benchmark Validation. Turkish-language benchmark rankings (FLORES-200, multilingual MMLU, multilingual MT-Bench) derived from published evaluations independent of the present wargame data. *GPT-4o provisionally classified; null re*

| Model | TR ML Benchmark Rank | Taxonomy | Obs. δ CoR | Consistency |
|---|---|---|---|---|
| Llama-4-Maverick | Mid | Type I | +0.800 | Consistent: low ML rank → susceptible |
| GPT-4o-2024-11-20 | High | Type II* | +0.130 | Consistent: high ML rank → invariant/null |
| Gemini-3.1-Pro | Very High | Type III | −0.750 | Consistent: highest ML rank → reversal |
| Mistral-Large-2512 | Mid-Low | Unstable | −0.600 | Partial: low rank → instability |
| DeepSeek-R1 | Mid-High (+CoT) | Type III | −0.860 | Consistent: CoT + mid-high rank |
| Qwen3.6-Plus | Mid | Null-Med. | −0.470 | Partial: mid rank → weak effect |

### *4.10 Aggregate Effects and Western Cluster Analysis*

The pooled Western cluster analysis (Hellenia + Gallica + Tarsia) on Coercive Rhetoric yields: EN $\mu = 0.661$, TR $\mu = 0.685$, $\delta = +0.089$ (Negligible), $p = .551$. The aggregate Shibboleth Effect across Western models is statistically indistinguishable from zero. This falsifies the strong version of the original hypothesis in favor of a model-specific, heterogeneous account.

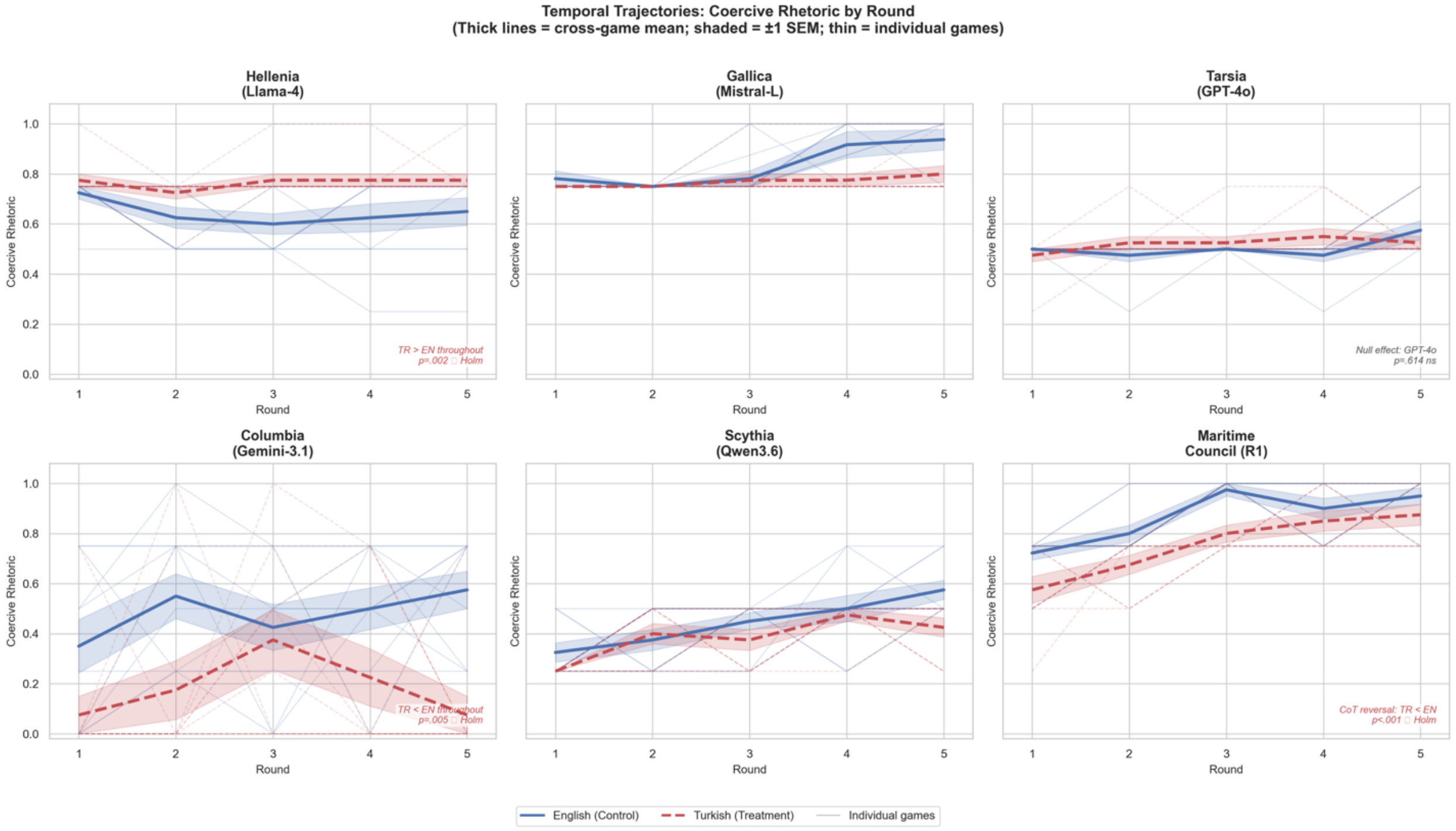

*Figure 2: Temporal Trajectories: Coercive Rhetoric by Round*

*Note: Thick lines = cross-game mean; shaded = ±1 SEM; thin = individual game observations. Panel annotations indicate Holm-corrected findings and null results*

## 5 Discussion

### *5.1 A Provisional Descriptive Taxonomy of Cross-Lingual Behavioral Stability*

We propose, as a post-hoc descriptive taxonomy awaiting prospective validation, a three-tier classification of cross-lingual behavioral stability. We use 'taxonomy' rather than 'typology' deliberately: the classifications rest on single-model observations and do not constitute an explanatory theoretical framework. External multilingual benchmark data (Table 6) furnishes non-circular construct validity for 4 of 6 classifications.

Type I (Shibboleth-Susceptible) encompasses models whose English-dominant RLHF alignment is insufficiently generalized to non-English contexts, producing coercion amplification under Turkish. Llama-4 is the present exemplar. Falsifiable prediction: Llama-4 should exhibit the same directional result in Arabic and Mandarin wargame arms.

Type II (Alignment-Invariant) encompasses models maintaining approximately equivalent behavioral profiles across English and Turkish. GPT-4o provisionally represents this type, though the null result is inconclusive at N = 10. Falsifiable prediction: GPT-4o should show null or negligible effects across additional language pairs. Replication at N = 25 would provide adequate power to distinguish Type II from Types I and III.

Type III (Reversal) encompasses models exhibiting reduced coercion under Turkish, via multilingual alignment (Gemini-3.1-Pro) or chain-of-thought institutional anchoring (DeepSeek-R1). For R1, the buffer is directly observed in captured reasoning chains; for Gemini, it is inferred from behavioral output and corroborated by external benchmark performance. Falsifiable prediction: both should exhibit the same directional result in Arabic and Mandarin arms.

### *5.2 Two Buffering Mechanisms*

Prior work hypothesized that chain-of-thought reasoning provides a structural buffer against the Shibboleth Effect. The present study furnishes the first direct evidence: DeepSeek-R1's captured reasoning chains explicitly invoke the UNCLOS mandate before generating public statements, functioning as a cognitive anchor. This mechanism is observed in the data, not inferred post hoc. The Gemini result establishes that CoT reasoning is not the only buffering pathway. The most parsimonious explanation—independently supported by external benchmark performance—is multilingual RLHF investment. These two mechanisms are not mutually exclusive and may be additive: reasoning-capable models with strong multilingual RLHF training should exhibit the most robust cross-lingual behavioral stability.

### *5.3 Algorithmic Constructivism: A Heuristic, Not a Theoretical Claim*

Constructivism as a theoretical lens for understanding how LLMs adopt geopolitical postures conditioned on linguistic discourse. The limits of this analogy warrant explicit acknowledgment.

Constructivism is a social theory of identity formation through intersubjective interaction—state identities are co-constituted through diplomatic practice(Wendt 1992, 1994, 1999). LLMs do not possess identities formed through interaction; they have frozen weight configurations that produce outputs. The Constructivist analogy maps the functional outcome (behavior conditioned on linguistically constituted discourse) onto the Constructivist account of how historical narrative shapes strategic posture, without claiming the underlying mechanism is intersubjective in Wendt's sense. We deploy Algorithmic Constructivism as an analogical heuristic to illuminate the pattern of results, not as a foundational theoretical claim.

### *5.4 Geopolitical Ramifications*

The heterogeneous Shibboleth Effect poses a more complex systemic risk than the original formulation suggested. It is no longer sufficient to ask whether a model exhibits cross-lingual bias; the policy-relevant questions concern what type of bias, in which direction, and with what magnitude. A diplomat querying Llama-4 in English receives cooperative, conflict-averse strategic analysis; the same query in Turkish may produce significantly more coercive framing. A diplomat querying Gemini-3.1-Pro in Turkish may receive significantly more cooperative advice than an English-language query would suggest—a reversal that could equally distort decision-making by under-representing conflict severity. Properly understood, the systemic risk is not that LLMs are uniformly hostile in non-English languages, but that the direction and magnitude of cross-lingual behavioral shift is model-specific and typically unknown to the end user.

### *5.5 Mechanism Underdetermination*

The present study measures the net behavioral consequence of language-arm assignment. Whether the mechanism is phonological token routing, corpus-content association, or their interaction cannot be separated within the current design, because language and the corpus that constitutes it are co-constituted. The Gallica partial test (Table 5) demonstrates that a pure corpus-content account is insufficient: the EU-proxy role is predicted to produce elevated coercion in Turkish but instead yields reduced coercion; the Turkey-proxy role (Tarsia) should benefit most from Turkish corpus alignment but shows a null effect. A future design capable of resolving this would require a third language arm using Turkish syntax with neutralized geopolitical valence, and counterbalanced role assignments across models.

### *5.6 Limitations and Future Work*

Role-language confounding is the primary limitation of this study. Because role-to-model assignments were not counterbalanced across conditions, role-specific associations in Turkish-language corpora may have interacted with the language manipulation in ways this design cannot fully separate. Resolving this issue will require a counterbalanced experimental design.

Human validation is also absent from the present dataset. The prior study's Cohen's $\kappa = 0.84$ was calculated on a different dataset, so the automated judge's cross-lingual construct validity has not yet been independently verified for the data used here.

The study is also limited to a single scenario and a single language pair. Future work should replicate the design across scenarios such as nuclear brinkmanship, economic statecraft, and humanitarian intervention, and should test additional languages including Arabic, Mandarin, Hindi, and Swahili, each of which draws on distinct geopolitical corpora.

Statistical power is another important limitation. With $N = 10$ per arm under a Holm-corrected $\alpha = .0042$ threshold, the study is adequately powered only to detect large effects when ceiling compression is absent. As a result, all null and non-significant findings should be treated as inconclusive. Increasing the sample size to $N = 25$–$30$ per arm would substantially improve power to detect medium-sized effects.

A further limitation is model-version substitution. Replacing GPT-5.5 with GPT-4o introduces a confound between model family and training-regime effects. Future factorial designs should explicitly separate architecture, training regime, and language arm so their individual contributions can be assessed more clearly.

## 6 Conclusion

This study empirically dismantles the monolithic formulation of the Shibboleth Effect while confirming its existence as a real, model-specific phenomenon. Across ten statistically independent game iterations per language arm, cross-lingual behavioral skew has been quantified with methodological rigor that prior single-run designs could not achieve.

The central finding is heterogeneity rather than uniformity. Llama-4 exhibits large coercion amplification under Turkish; Gemini-3.1-Pro exhibits an equally large cooperative amplification; GPT-4o is null and inconclusive at $N = 10$; DeepSeek-R1 exhibits a strong reversal supported by direct chain-of-thought evidence. The pooled Western cluster effect is statistically zero. The Shibboleth Effect is not a property of Western-origin models as a categorical class; it is a property of specific training regimes and architectural configurations.

We propose a provisional descriptive taxonomy—Type I (Susceptible), Type II (Invariant), Type III (Reversal)—corroborated by external multilingual benchmark data and accompanied by falsifiable cross-scenario predictions. Two structurally distinct buffering mechanisms are identified: chain-of-thought institutional anchoring (directly evidenced in DeepSeek-R1's captured reasoning) and multilingual RLHF alignment (inferred for Gemini-3.1-Pro). For decision-makers integrating LLMs into crisis management, the policy implication is precise: cross-lingual behavioral stability is not a given, its direction cannot be assumed, and it varies across model versions within the same model family. Robust cross-lingual alignment evaluation—conducted under adversarial friction, across multiple independent replications, with explicit Holm-corrected hypothesis testing—must become a standard component of LLM safety certification for high-stakes deployment contexts.

**Statements and Declarations**

**Competing Interests:** The authors have no relevant financial or non-financial interests to disclose.

**Funding:** No funding was received to assist with the preparation of this manuscript.

**Ethics Approval:** This study involved no human participants. The experimental subjects are large language model systems; no ethics committee approval was required.

**Data Availability Statement:** The simulation logs (20 JSONL files, 10 EN + 10 TR), the scored dataset (evaluation_results.csv), the scenario briefings (Cerulean_Sea_Crisis_EN.json, Cerulean_Sea_Crisis_TR.json), and all analysis scripts (diplomacy_engine.py, judge_evaluator.py, evaluation.py, visualization.py) are deposited in the following project data repository: Mehmetcik, Hakan (2026). Replication Data for The Shibboleth Effect: Auditing the Cross-Lingual Distributional Skew of Large Language Models. figshare. Dataset. https://doi.org/10.6084/m9.figshare.32389938.v1

Note: The OPENROUTER_API_KEY required to re-run the simulation is not deposited; researchers must obtain their own key from openrouter.ai. The simulation can be reproduced deterministically using the logged BASE_SEED = 20260521 and the seed formula documented in Sect. 3.3.

**AI Use Disclosure:** This study uses large language models in two experimental capacities: (1) as research subjects (the six frontier models constituting the flagship ecology); and (2) as an automated scoring instrument (gpt-4o-2024-05-13 as zero-shot judge evaluator, Temperature = 0.0). These uses are described in full in Sect. 3.5. No AI tool was used for manuscript drafting, copy editing, or translation.

## References

Banerjee S, Agarwal A, Singh E (2024) The Vulnerability of Language Model Benchmarks: Do They Accurately Reflect True LLM Performance? https://doi.org/10.48550/ARXIV.2412.03597

Dahlgren Lindström A, Methnani L, Krause L, et al (2025) Helpful, harmless, honest? Sociotechnical limits of AI alignment and safety through Reinforcement Learning from Human Feedback. Ethics Inf Technol 27:28. https://doi.org/10.1007/s10676-025-09837-2

Giacalone M (2026) Discursive behavior of generative language models in geopolitical and humanitarian contexts. Discov Artif Intell 6:230. https://doi.org/10.1007/s44163-026-00965-2

Goldstein JA, Sastry G, Musser M, et al (2023) Generative language models and automated influence operations: Emerging threats and potential mitigations. arXiv preprint arXiv:230104246 1:

Guey W, Bougault P, Moura VD de, et al (2025) Mapping Geopolitical Bias in 11 Large Language Models: A Bilingual, Dual-Framing Analysis of U.S.-China Tensions

Hua W, Fan L, Li L, et al (2024) War and Peace (WarAgent): Large Language Model-based Multi-Agent Simulation of World Wars

Jonnala S, Swamy B, Thomas NM (2025) Geopolitical Bias in Sovereign Large Language Models: A Comparative Mixed-Methods Study. J Res Innov Technol 4:173–192

Kerche FW, Zook M, Graham M (2026) The silicon gaze: A typology of biases and inequality in LLMs through the lens of place. Platforms & Society 3:29768624251408919. https://doi.org/10.1177/29768624251408919

Kotarski K, Mudrić MM, Dragović F (2026) The Double-Edged Algorithm: Addressing the Dual-Use of AI Technologies in the Age of Great Power Competition Through the Lens of EU. In: Vaseashta A, Rocha A (eds) Spectrum of Dual-Use Technologies: Unforeseen Risks Versus Returns. Springer Nature Switzerland, Cham, pp 205–227

Leite JA, Arora A, Gargova S, et al (2025) A Multilingual, Large-Scale Study of the Interplay between LLM Safeguards, Personalisation, and Disinformation

Li B, Haider S, Callison-Burch C (2024a) This Land is Your, My Land: Evaluating Geopolitical Bias in Language Models through Territorial Disputes. In: Proceedings of the 2024 Conference of the North American Chapter of the Association for Computational Linguistics: Human Language Technologies (Volume 1: Long Papers). Association for Computational Linguistics, Mexico City, Mexico, pp 3855–3871

Li Y, Huang Y, Wang H, et al (2024b) Evaluating Large Language Models with Psychometrics

Matlin G, Mahajan P, Song I, et al (2025) Shall We Play a Game? Language Models for Open-ended Wargames

Miller A (2026) AI multi-agent reinforcement learning (MARL) for conflict resolution and forecasting in international relations policy. International Journal of Computing and Artificial Intelligence 7:32–37. https://doi.org/10.33545/27076571.2026.v7.i2a.261

Mouakher A, Morgado N, Ftouhi F (2026) LLM4Geopolitics: A Framework Leveraging Large Language Models for Predicting Geopolitical Events. Expert Systems 43:e70258. https://doi.org/10.1111/exsy.70258

Olivieri AF, Guadagno RE, Solari S, Russo E (2026) EWACS as a Backbone for Wargaming and Decision Support in the Information Environment

Pacheco AGC, Cavalini A, Comarela G (2026) Echoes of power: investigating geopolitical bias in US and China large language models. Humanit Soc Sci Commun 13:675. https://doi.org/10.1057/s41599-026-06577-6

Rivera J-P, Mukobi G, Reuel A, et al (2024) Escalation Risks from Language Models in Military and Diplomatic Decision-Making. In: The 2024 ACM Conference on Fairness Accountability and Transparency. ACM, Rio de Janeiro Brazil, pp 836–898

Röttger P, Hofmann V, Pyatkin V, et al (2024) Political Compass or Spinning Arrow? Towards More Meaningful Evaluations for Values and Opinions in Large Language Models. arXiv

Ryström JH, Kirk HR, Hale SA (2025) Multilingual!= multicultural: Evaluating gaps between multilingual capabilities and cultural alignment in llms. In: Proceedings of Interdisciplinary Workshop on Observations of Misunderstood, Misguided and Malicious Use of Language Models. pp 74–85

Salnikov M, Korzh D, Lazichny I, et al (2025) Geopolitical biases in LLMs: what are the “good” and the “bad” countries according to contemporary language models

Shankar H, P VS, Margapuri S, et al (2026) Mind the Gap: Pitfalls of LLM Alignment with Asian Public Opinion

Shen L, Tan W, Chen S, et al (2024) The Language Barrier: Dissecting Safety Challenges of LLMs in Multilingual Contexts. arXiv

Shrivastava A, Hullman J, Lamparth M (2024) Measuring Free-Form Decision-Making Inconsistency of Language Models in Military Crisis Simulations

Smirnov O (2026) The Language You Ask In: Language-Conditioned Ideological Divergence in LLM Analysis of Contested Political Documents. https://doi.org/10.48550/ARXIV.2601.12164

Solopova V, Skorik V, Tereshchenko M, et al (2026) LLMs as Strategic Actors: Behavioral Alignment, Risk Calibration, and Argumentation Framing in Geopolitical Simulations

Sornette D, Lera SC, Wu K (2026) Why AI Alignment Failure Is Structural: Learned Human Interaction Structures and AGI as an Endogenous Evolutionary Shock. In: arXiv.org. https://arxiv.org/abs/2601.08673v1. Accessed 25 May 2026

Srivastava S, Janardhan K, Jauhari S (2026) A Systematic Review of Algorithmic Red Teaming Methodologies for Assurance and Security of AI Applications. https://doi.org/10.48550/ARXIV.2602.21267

Sukiennik N, Gao C, Xu F, Li Y (2025) An Evaluation of Cultural Value Alignment in LLM

Tao Y, Viberg O, Baker RS, Kizilcec RF (2024) Cultural bias and cultural alignment of large language models. PNAS nexus 3:pgae346

Triantafyllopoulos L, Paxinou E, Tzanoulinou D, et al (2026) The value alignment problem in advisory AI: a systematic literature review. AI Ethics 6:147. https://doi.org/10.1007/s43681-026-01015-4

Weller D, Meltschack M, Schwindling D (2024) Leveraging Large Language Models for Enhanced Wargaming in Multi-Domain Operations

Wendt A (1992) Anarchy is what states make of it: the social construction of power politics. International organization 46:391–425

Wendt A (1994) Collective identity formation and the international state. American political science review 88:384–396

Wendt A (1999) Social theory of international politics. Cambridge university press

Wu J, Chen S, Chen P, et al (2026) A resource-efficient framework for cultural alignment in large language models (LLMs): The Chinese context. Design and Artificial Intelligence 2:100063. https://doi.org/10.1016/j.daai.2026.100063

Wu M-C, Chin S-C, Wood T, et al (2025) Incorporating diverse perspectives in cultural alignment: Survey of evaluation benchmarks through a three-dimensional framework. In: Proceedings of the 2025 Conference on Empirical Methods in Natural Language Processing. pp 17037–17072

Ye H, Jin J, Xie Y, et al (2025) Large Language Model Psychometrics: A Systematic Review of Evaluation, Validation, and Enhancement. https://doi.org/10.48550/ARXIV.2505.08245

Yong Z-X, Menghini C, Bach SH (2023) Low-Resource Languages Jailbreak GPT-4. https://doi.org/10.48550/ARXIV.2310.02446

Yu S, Choi J, Kim Y (2025a) Delving into Multilingual Ethical Bias: The MSQAD with Statistical Hypothesis Tests for Large Language Models. arXiv

Yu S, Cui L, Liu S, Huang K (2025b) Stable and Expert-Aligned Evaluation of Wargaming Strategies via Optimized LLM Scoring Agents. In: 2025 6th International Conference on Machine Learning and Computer Application (ICMLCA). pp 1013–1017

Zahraei PS, Asgari E (2025) I Am Aligned, But With Whom? MENA Values Benchmark for Evaluating Cultural Alignment and Multilingual Bias in LLMs

Zhou D, Zhang Y (2024) Political biases and inconsistencies in bilingual GPT models—the cases of the U.S. and China. Sci Rep 14:25048. https://doi.org/10.1038/s41598-024-76395-w

---

[i] The prior version of this manuscript by the same author(s) is cited to establish the study's place within the cumulative research program. The citation has been anonymized for blind review in accordance with journal's double-anonymous peer review policy. Full bibliographic details will be restored upon acceptance.